\documentclass[conference]{IEEEtran}
\usepackage{cite}
\usepackage{amsmath,amssymb,amsfonts}
\usepackage{algorithm}
\usepackage[noend]{algorithmic}
\usepackage{graphicx}
\usepackage{textcomp}
\usepackage{xcolor}
\usepackage{multirow}
\usepackage{bbm}
\def\BibTeX{{\rm B\kern-.05em{\sc i\kern-.025em b}\kern-.08em T\kern-.1667em\lower.7ex\hbox{E}\kern-.125emX}}
    
\begin{document}

\title{Real-time Traffic Classification\\for 5G NSA Encrypted Data Flows\\With Physical Channel Records}
 
\author{\IEEEauthorblockN{Xiao Fei}
\IEEEauthorblockA{\textit{Shanghai Jiao Tong University} \\
\textit{colinfx@sjtu.edu.cn}}
\and
\IEEEauthorblockN{Philippe Martins*}
\IEEEauthorblockA{\textit{Telecom Paris, Institut Polytechnique de Paris} \\
\textit{martins@telecom-paris.fr}}
\and
\IEEEauthorblockN{Jialiang Lu*}
\IEEEauthorblockA{\textit{Shanghai Jiao Tong University} \\
\textit{jialiang.lu@sjtu.edu.cn}}
}

\maketitle

\begin{abstract}
The classification of fifth-generation New-Radio (5G-NR) mobile network traffic is an emerging topic in the field of telecommunications. It can be utilized for quality of service (QoS) management and dynamic resource allocation. However, traditional approaches such as Deep Packet Inspection (DPI) can not be directly applied to encrypted data flows. Therefore, new real-time encrypted traffic classification algorithms need to be investigated to handle dynamic transmission. In this study, we examine the real-time encrypted 5G Non-Standalone (NSA) application-level traffic classification using physical channel records. Due to the vastness of their features, decision-tree-based gradient boosting algorithms are a viable approach for classification. We generate a noise-limited 5G NSA trace dataset with traffic from multiple applications. We develop a new pipeline to convert sequences of physical channel records into numerical vectors. A set of machine learning models are tested,  and we propose our solution based on Light Gradient Boosting Machine (LGBM) due to its advantages in fast parallel training and low computational burden in practical scenarios. Our experiments demonstrate that our algorithm can achieve 95\% accuracy on the classification task with a state-of-the-art response time as quick as 10ms. 
\end{abstract}

\section{Introduction}\label{sec:intro}
Mobile network real-time traffic classification through the analysis of uplink and downlink data streams, without direct access to the encrypted user data, has been widely studied \cite{bib:tc1}. Such traffic identification is considered to be of great value for radio resources allocation optimisation, QoS evaluation, malware traffic detection, Mobile Network Operator (MNO) policy management and network slicing \cite{bib:policy}. 

There are two major challenges associated with this task. First, the encryption of data flow renders traditional approaches that rely on sensitive user data inapplicable, leaving only indirect features available for classification. Second, the rapid increase in transmission rates and the vast expansion of system capacity impose stringent demands on the response time of identification \cite{bib:realtime}. 

Traditionally, the Internet Protocol (IP) address and the port have been used as simple and direct indicators for traffic classification \cite{bib:port}. However, many ports remain unregulated, and the same port may correspond to different services in varying scenarios. Furthermore, this information cannot be accessed from the radio interface as flows are encrypted at Packet Data Convergence Protocol (PDCP) layer. 

As another traditional approach, DPI involves directly reading the core data packets in transmission and classifying them based on plaintext content fragments \cite{bib:dpi}. However, it is highly complex and slow to compute, and its performance in encrypted scenarios has proven to be unsatisfactory. 

To tackle with the challenge of encrypted data flows, statistical classification methods using various features with learning algorithms are proposed. Existing studies can be categorized into two approaches based on different sources. The first category extracts characteristic features primarily from L3 IP data packets \cite{bib:cnn,bib:ae,bib:gcnn,bib:mm}. Many studies achieved exceptional results in application-level identification with various deep learning (DL) frameworks, most of which are Convolutional Neural Network (CNN), Recurrent Neural Network (RNN), Auto Encoder, Graph CNN (GCNN) and multimodal combinations of them. 

In contrast, the second category uses time-series features of signal waves, such as the spectrum obtained by Fourier transform and the variation in amplitude as a function of time \cite{bib:cnn2,bib:lstm}. CNN and Long Short Term Memory (LSTM) frameworks are most commonly employed to extract time-series characteristics of the trace. However, such methods are often susceptible to noise and interference and can only identify information strongly associated with signals, such as modulation options and protocol types. 

However, the problem of real-time traffic classification remains unsolved. Both statistical classification approaches failed to meet such need, as the first requires a lengthy sequence of IP packets, and the second relies on signal waves spanning several seconds. 

\begin{table*}[t]
\centering
\caption{Comparison of Traffic Classification Frameworks}
\label{tab:rw}
\begin{tabular}{ccccccc}
\hline
\multicolumn{1}{|c|}{Approach} & \multicolumn{1}{c|}{Framework} & \multicolumn{1}{c|}{Features} & \multicolumn{1}{c|}{Target} & \multicolumn{1}{c|}{Consistency} & \multicolumn{1}{c|}{Encrypted} & \multicolumn{1}{c|}{Real-time} \\ \hline
\multicolumn{1}{|c|}{Port Inspection} & \multicolumn{1}{c|}{Look-up Table \cite{bib:port}} & \multicolumn{1}{c|}{\begin{tabular}[c]{@{}c@{}}IP address\\ Port number\end{tabular}} & \multicolumn{1}{c|}{Applications} & \multicolumn{1}{c|}{Poor} & \multicolumn{1}{c|}{Yes} & \multicolumn{1}{c|}{Yes} \\ \hline
\multicolumn{1}{|c|}{DPI} & \multicolumn{1}{c|}{Learning Algorithm \cite{bib:dpi}} & \multicolumn{1}{c|}{IP packet plaintext contents} & \multicolumn{1}{c|}{Applications} & \multicolumn{1}{c|}{Good} & \multicolumn{1}{c|}{No} & \multicolumn{1}{c|}{No} \\ \hline
\multicolumn{1}{|c|}{\multirow{2}{*}{\begin{tabular}[c]{@{}c@{}}Statistical\\ Classification\end{tabular}}} & \multicolumn{1}{c|}{\begin{tabular}[c]{@{}c@{}}CNN \cite{bib:cnn}\\ Auto Encoder \cite{bib:ae}\\ GCNN \cite{bib:gcnn}\\ Multimodal \cite{bib:mm}\end{tabular}} & \multicolumn{1}{c|}{\begin{tabular}[c]{@{}c@{}}IP packet length\\ Inter-arrival time\\ Initial bytes\\ Other packet statistics\end{tabular}} & \multicolumn{1}{c|}{Applications} & \multicolumn{1}{c|}{Good} & \multicolumn{1}{c|}{\multirow{2}{*}{Yes}} & \multicolumn{1}{c|}{\multirow{2}{*}{No}} \\ \cline{2-5}
\multicolumn{1}{|c|}{} & \multicolumn{1}{c|}{\begin{tabular}[c]{@{}c@{}}CNN \cite{bib:cnn2}\\ LSTM \cite{bib:lstm}\end{tabular}} & \multicolumn{1}{c|}{\begin{tabular}[c]{@{}c@{}}Spectrum\\ Variation in amplitude\end{tabular}} & \multicolumn{1}{c|}{\begin{tabular}[c]{@{}c@{}}Modulations\\ Protocols\end{tabular}} & \multicolumn{1}{c|}{Poor} & \multicolumn{1}{c|}{} & \multicolumn{1}{c|}{} \\ \hline
\multicolumn{1}{|c|}{Our approach} & \multicolumn{1}{c|}{\begin{tabular}[c]{@{}c@{}}Decision-tree-based\\ Gradient Boosting\end{tabular}} & \multicolumn{1}{c|}{Physical channel features} & \multicolumn{1}{c|}{Applications} & \multicolumn{1}{c|}{Good} & \multicolumn{1}{c|}{Yes} & \multicolumn{1}{c|}{Yes} \\ \hline
\\
\end{tabular}
\end{table*}

To address these two major challenges, we propose a novel approach that utilizes sequence of 5G physical channel records. The high-density features of it provide sufficient information for identification in a very short period of time. We design a new pipeline to process sequence of records in time-frequency domain to formatted data for any downstream tasks without delay. Additionally, our decision-tree-based gradient boosting framework excels at extracting pertinent information from complex features and imposes lighter computational burden compared to large neural networks. This makes application-level, real-time traffic classification possible. The Table \ref{tab:rw} compares our algorithm with existing state-of-the-art solutions. 

To simplify trivial details without losing the universality, we conducted experiments under the noise-limited 5G NSA network framework with one user device, performing at most one activity at a time. 

In the rest part of this paper, Section \ref{sec:pre} provides an overview the 5G architecture and the decision-tree-based gradient boosting algorithms, while our approach is elaborated in Section \ref{sec:metho}. Experimental results are presented in Section \ref{sec:res} and conclusions are drawn in Section \ref{sec:con}. 

\section{Preliminaries}\label{sec:pre}

\begin{figure}[htbp]
    \centering
    \includegraphics[width=\columnwidth]{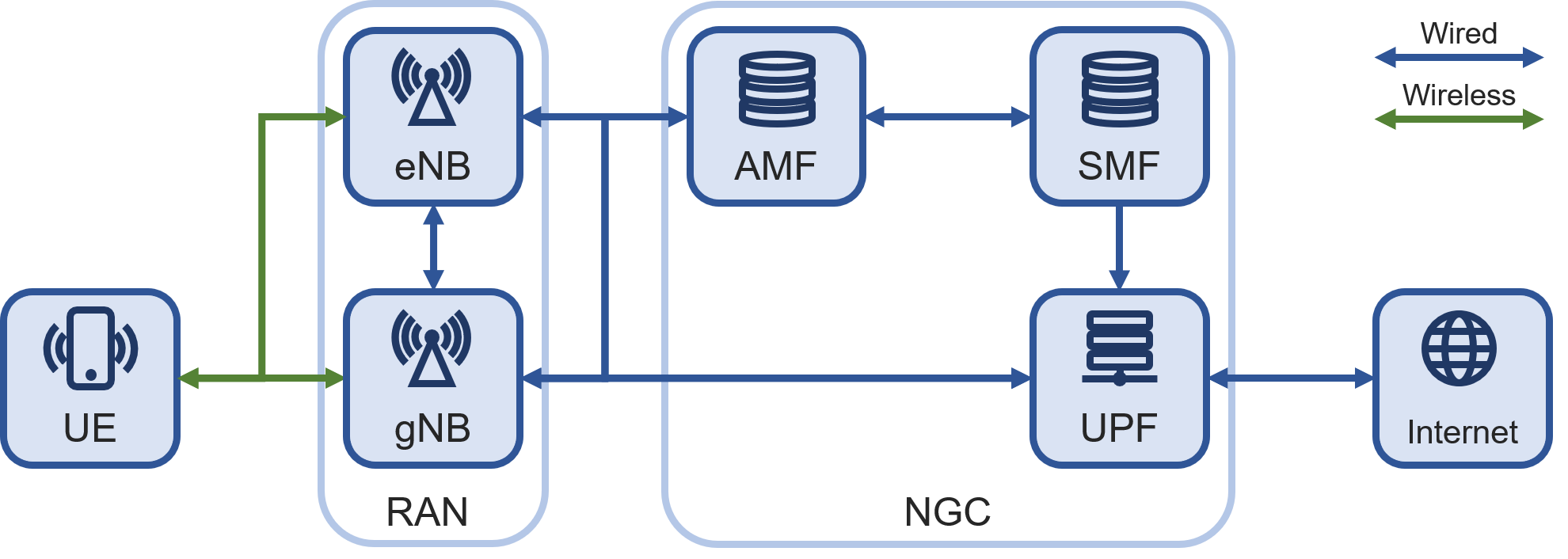}
    \caption{5G NSA Network Architecture (Option 4)}
    \label{fig:nsafull}
\end{figure}

The 5G-NR system is composed of three parts: the User Equipment (UE), typically smartphones, the Radio Access Network (RAN) which allocates radio resources to UEs, and the Next Generation Core (NGC), which is responsible for communication with the Internet and managing the entire system. The air interface between UE and RAN is wireless while all other connections are wired, as shown in Figure \ref{fig:nsafull}. 

Both 5G Non-Standalone (NSA) and Standalone (SA) architectures are available. SA employs NGC as core network and gNodeB (gNB) as RAN, but incurs high cost in hardware upgrades. In contrast, NSA serves as an intermediate transition. It retains the 4G Long Term Evolution (LTE) eNodeB (eNB) while adding a gNB in parallel and transmits data on both cells $cell_{lte}, cell_{nr}$. Some NSA options use the existing 4G Evolved Packet Core (EPC) instead of the NGC. 

\begin{figure}[htbp]
    \centering
    \includegraphics[width=\columnwidth]{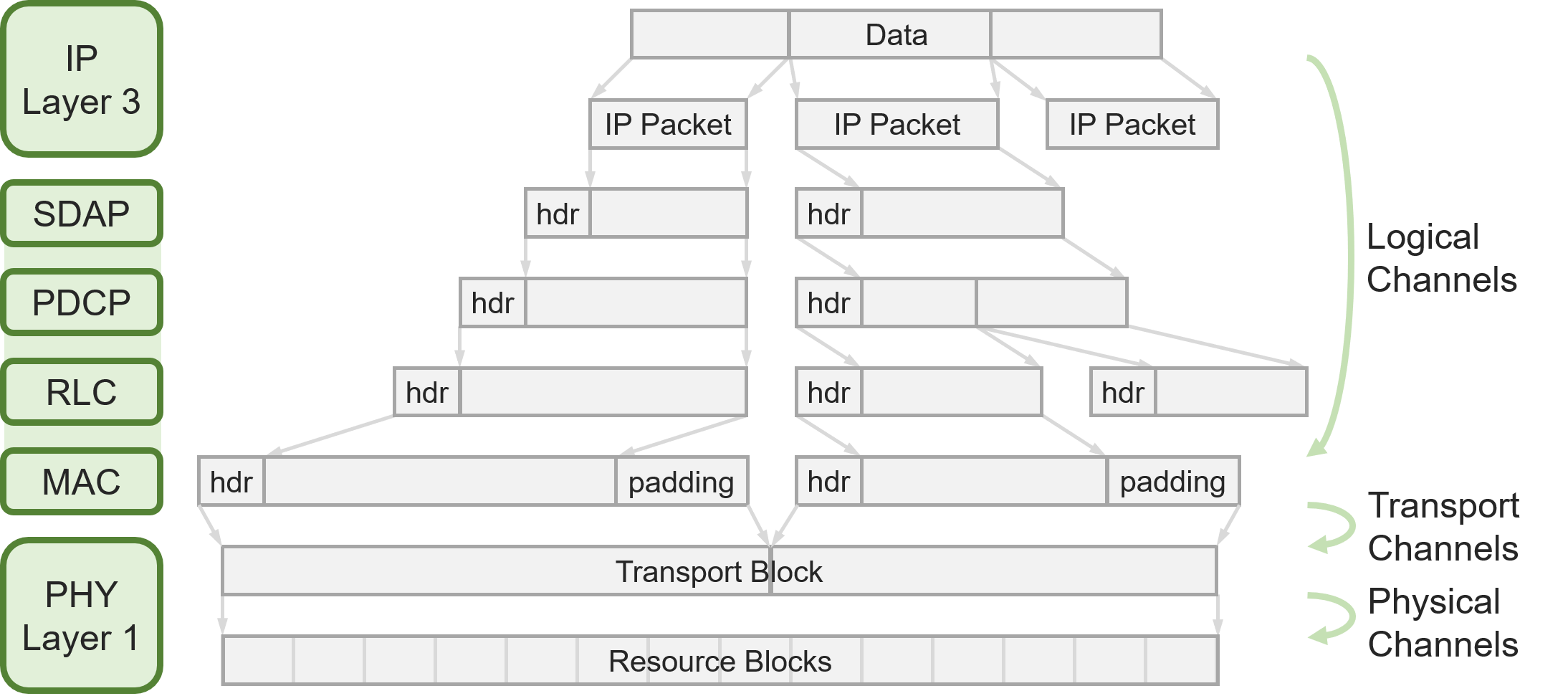}
    \caption{5G NR Protocol Stack and Data Flow}
    \label{fig:dataflow}
\end{figure}

Meanwhile, the air interface follows a protocol stack that converts data streams back and forth to electromagnetic waves at the transmitter and receiver. As shown in Figure \ref{fig:dataflow}, raw data from L3 IP layer is segmented into a series of IP packets, which are then processed by layer 2 stack into Transport Blocks (TB). They are taken by L1 Physical Layer for transmission by electromagnetic waves. 

During this process, user data and control information are passed through different channels at each layer following hierarchical rules. Physical channels, located at the bottom of this protocol stack, are responsible for the allocation of frequency and time resources to data. 

\begin{figure}[b!]
    \centering
    \includegraphics[width=\columnwidth]{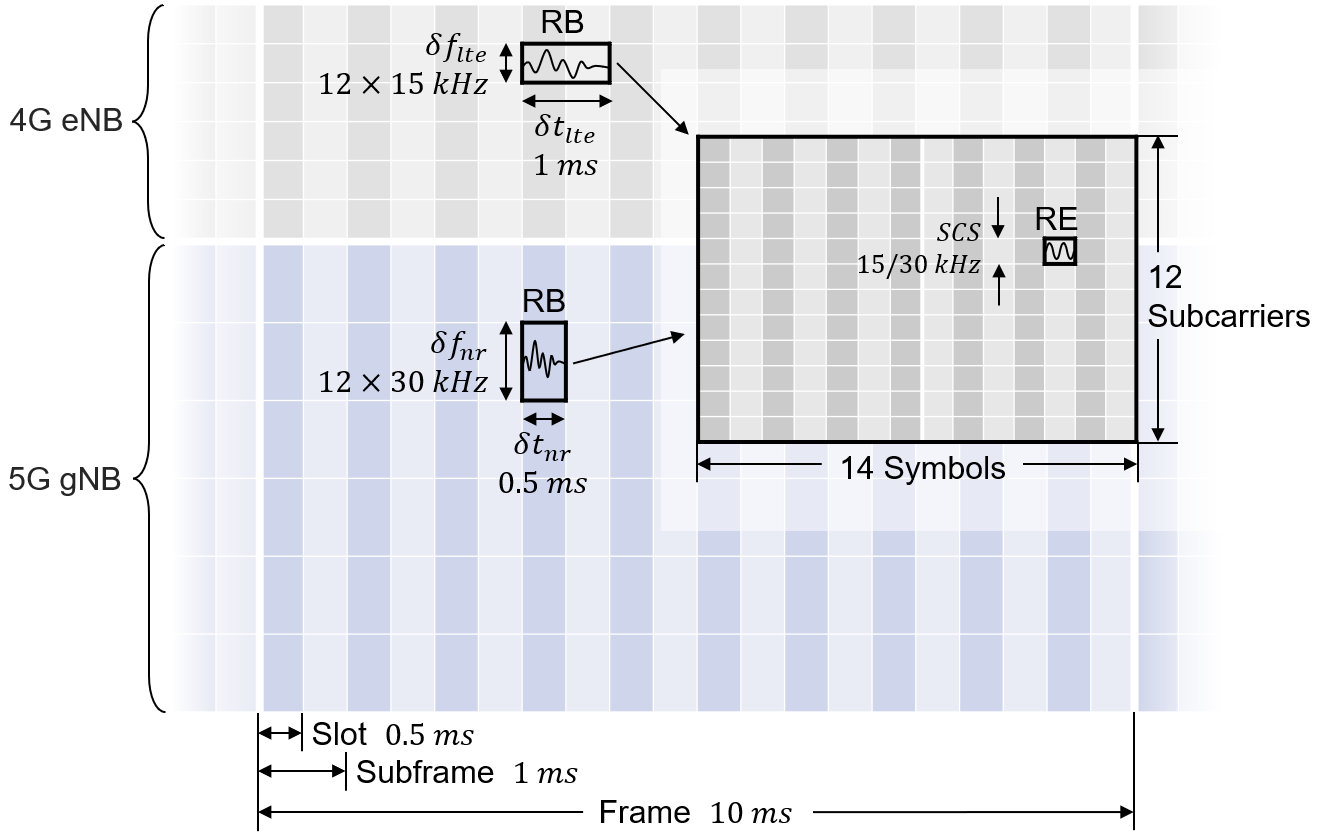}
    \caption{5G NR NSA Frame: Frequency and Time Resource Allocation}
    \label{fig:allocation}
\end{figure}

Thanks to orthogonal frequency-division multiple access (OFDMA) technology, it is possible to segment the entire bandwidth into subcarriers without interference between them. The time axis is segmented into symbols, so that each Resource Element $RE_{i,j}|_{i\in\{1,...,sc_{num}\},j\in\mathbb{N}}$, corresponding to a specific frequency $f_i$ and a specific short period of time $[t_j,t_{j+1}]$, carries a single frequency electromagnetic wave for transmission. 

On the time axis, both 4G LTE and 5G with $\mu=1$ numerology configurations are designed with a frame of duration of $10 ms$, which can be segmented into subframes of $1 ms$, each consisting of two slots of $0.5 ms$. In 4G LTE, a Physical Resource Block (PRB) is composed of 12 subcarriers, each with a subcarrier spacing (SCS) of 15 kHz $\delta f_{lte}=12\times15 kHz$, times 7 symbols in a slot. Two adjacent PRBs sharing a subframe are combined as one Resource Block (RB) $\delta t_{lte} = 2\times 0.5 ms$, i.e. an RB contains $14\times 12$ REs. Similarly, 12 subcarriers times 14 symbols is an RB in 5G NR, but with 30 kHz as the SCS $\delta f_{nr}=12\times 30 kHz$, and of duration $\delta t_{nr}=0.5 ms$. The smallest unit for allocation in different channels is always the RB, which carries a composite wave signal. 

\section{Methodology}\label{sec:metho}

\subsection{Dataset Generation}

Simulations were conducted under NSA architecture for further extrapolation, as resource allocation maps for both LTE and SA are subsets of NSA. The problem was also simplified to the noise-limited case, considering the development of government radio-frequency management and of frequency-division technology. 

Our dataset was generated by the laboratory of INFRES\footnote{Department of Computer Science and Networks, Telecom-Paris, https://www.telecom-paris.fr/fr/lecole/departements-enseignement-recherche/informatique-reseaux}. A NSA architecture was set up in a Faraday cage with a single smartphone as terminal device to eliminate interference. Only one application was used at a time, as is most often the case in real-life scenarios, and traces are labelled according to different periods of time.  

Highly informative L1 Physical channel records were used because they are designed to accumulate and inherit service-specific information from upper channels, containing detailed transmission characteristics. Meanwhile, since large L3 data packets are segmented into small L1 resource blocks, any changes in the data stream will be reflected more quickly in the physical channel. 

However, the air interface transmitted numerous physical channels for various purposes, resulting in a vast set of features that were beyond the processing capabilities of machine learning models. Thus, we only kept important physical channels to reduce computation time, as list in Table \ref{tab:channels}. PDSCH and PUSCH\footnote{Physical Downlink/Uplink Shared Channel} contain detailed characteristics of user data, allowing us to overcome the encryption of the IP packets and obtain more implicit information. PDCCH, PUCCH\footnote{Physical Downlink/Uplink Control Channel}, SRS\footnote{Sounding Reference Signal} and PHICH\footnote{Physical HARQ (Hybrid Automatic Repeat Request) Indicator Channel} transmit control information including resource allocation, radio environment and network load. Since different traffic may have similar uplink or downlink data streams, both directions should be considered comprehensively.

\begin{table}[h]
\centering
\caption{Physical Channels Selected for Traffic Classification}
\label{tab:channels}
\begin{tabular}{|c|l|}
\hline
Channel & \multicolumn{1}{c|}{Carried Information} \\ \hline
PDSCH & Downlink user data and UE demodulation information \\ \hline
PUSCH & Uplink user data and RAN demodulation information \\ \hline
PDCCH & Downlink control information \\ \hline
PUCCH & Uplink control information \\ \hline
SRS & Uplink complementary demodulation information \\ \hline
PHICH & Uplink control information (for LTE cell only) \\ \hline
\end{tabular}
\end{table}

\subsection{Processing Pipeline}

We merged RB segments on same $cell$ in one $\delta t_{cell}$ allocated to same physical channel into a single record with all detailed features. Thus, every time unit $\delta t_{cell}$ had at most one record for each channel in that cell. 

Despite the inclusion of features from only six different physical channels, many of them appeared infrequently in records and were still too numerous for machine learning models to process effectively. We selected over 60 features based on implication analysis to further improve response time, with most important ones listed in Table \ref{tab:features}. Features such as $tb\_len$ (TB length) and $prb$ (PRB allocation) are directly related to the user data throughput, but also influenced by environment of signal transmission. Therefore, many characteristics related to channel quality must be considered as well, $epre$ (Energy per RE), $snr$ (Signal-to-Noise Ratio) and $harq$ (HARQ indicator) for example.  

\begin{table}[h]
\centering
\caption{Key Features Selected for Traffic Classification}
\label{tab:features}
\begin{tabular}{|c|c|l|}
\hline
Category & Variable & \multicolumn{1}{c|}{Representation} \\ \hline
\multirow{2}{*}{\begin{tabular}[c]{@{}c@{}}User data\\ throughput\end{tabular}} & $tb\_len$ & Quantity of data \\ \cline{2-3} 
 & $prb$ & Allocated time-frequency resource \\ \hline
\multirow{3}{*}{\begin{tabular}[c]{@{}c@{}}Transmission\\ environment\end{tabular}} & $epre$ & Efficiency of channel \\ \cline{2-3} 
 & $snr$ & Quality of channel \\ \cline{2-3} 
 & $harq$ & Quality of transmission \\ \hline
\end{tabular}
\end{table}

\begin{figure}[htbp]
    \centering
    \includegraphics[width=\columnwidth]{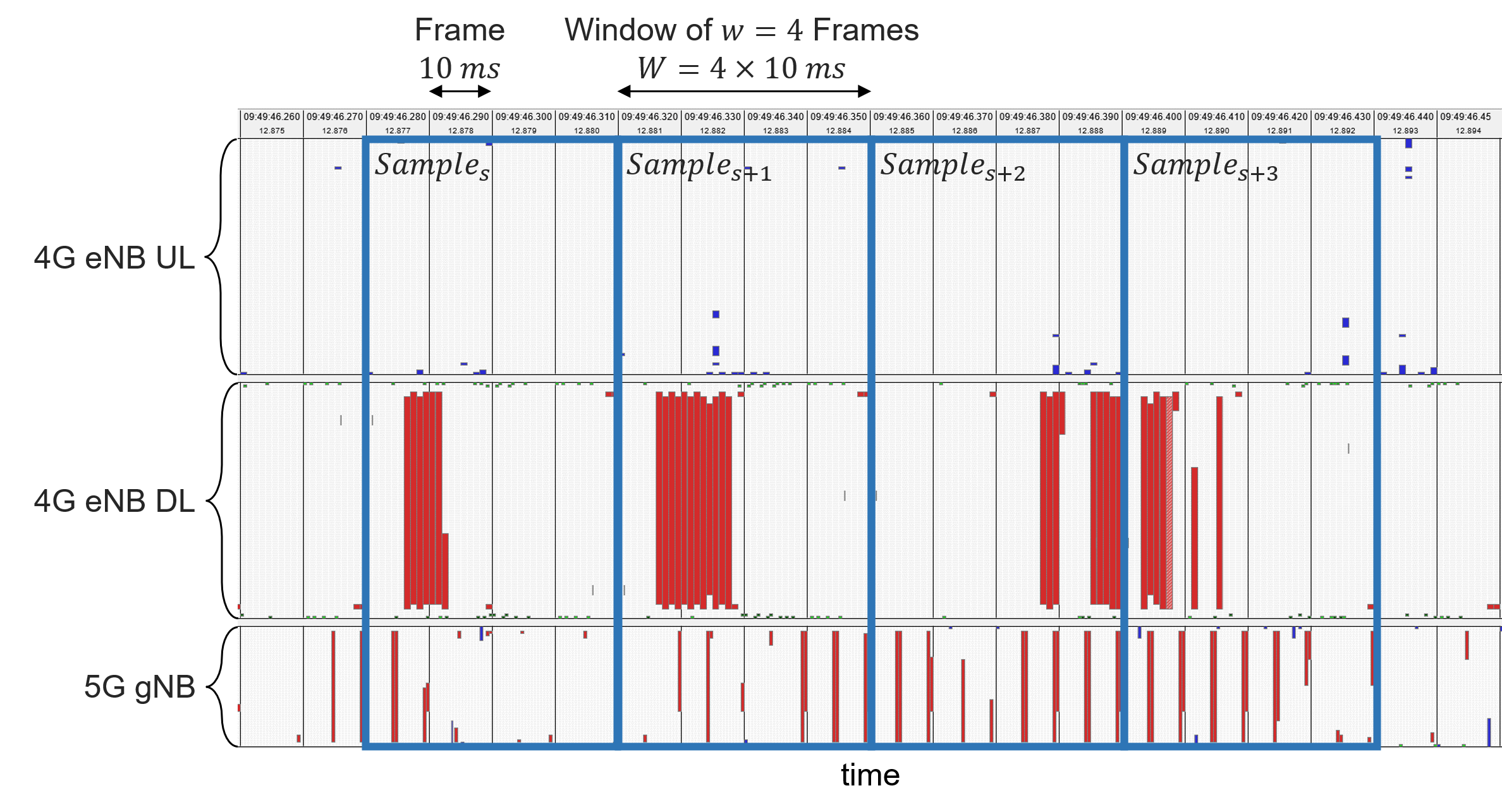}
    \caption{Extraction of Sample Vector From RB Scheme With Sliding Window}
    \label{fig:rbscheme}
\end{figure}

A sliding window with a duration of $W = 10w\, ms$ covering $w\in\mathbb{N^*}$ frames was applied, as illustrated in Figure \ref{fig:rbscheme}. Each time it captured one or multiple entire frames $(frame_f)_{f\in\{sw,...,sw+w-1\}}$, it was considered as one observation sample $sample_s$ to ensure that there were enough records in each sample. 

\begin{figure}[b!]
    \centering
    \includegraphics[width=\columnwidth]{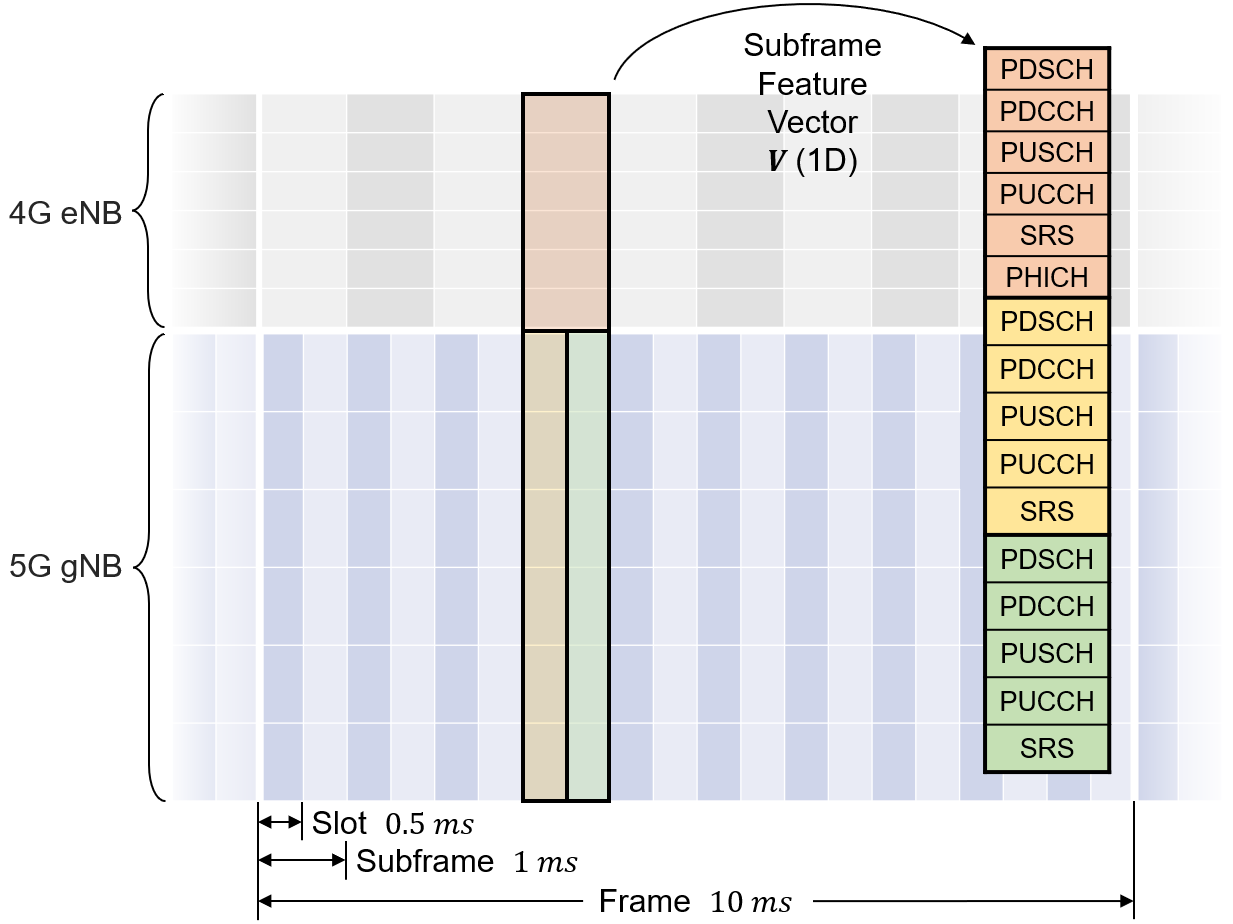}
    \caption{Forming of Feature Vector in 5G NSA Hybrid RB Allocation}
    \label{fig:featurevector}
\end{figure}

As many RBs remained vacant and the allocation schemes of different channels were irregular, we extracted a feature vector $\mathbf{V}$ for each subframe as illustrated in Figure \ref{fig:featurevector}. Each subframe $sf$ contains one time unit $[t_{sf}, t_{sf}+\delta t_{lte}]$ of 4G LTE and two time units $[t_{sf},t_{sf}+\delta t_{nr}],[t_{sf}+\delta t_{nr}, t_{sf}+2\delta t_{nr}]$ of 5G NR. 

As displayed in Algorithm \ref{algo:v}, if the channel record existed in $cell$, the extracted features were stored in the corresponding position. If not, zero padding was used to produce a feature vector of consistent length. Finally, feature vectors of subframes in the window were concatenated into a one- or two-dimensional array depending on the needs of downstream calculations.

\begin{algorithm}[t!]
\caption{Feature-vector $\mathbf{V}(sf)$}
\label{algo:v}
\begin{algorithmic}[1]
\renewcommand{\algorithmicrequire}{\textbf{Input:}}
\renewcommand{\algorithmicensure}{\textbf{Output:}}
\REQUIRE Records in one subframe $sf$
\ENSURE  Feature vector of the subframe $\mathbf{V}_{sf}$
\STATE $\mathbf{V} \leftarrow [\, ]$
\FOR {$cell\in$ [$cell_{lte}$, $cell_{nr,1}$, $cell_{nr,2}$]}
    \FOR {$channel \in channels_{cell}$}
        \IF {$\exists record_{channel} \in cell$}
            \FOR {$feature \in features_{channel}$}
                \STATE $\mathbf{V}\leftarrow$ concat($\mathbf{V}$, $feature$)
            \ENDFOR
        \ELSE
            \STATE $\mathbf{V}\leftarrow$ concat($\mathbf{V}$, $ [0]\times \#features_{channel}$)
        \ENDIF
    \ENDFOR
\ENDFOR
\RETURN $\mathbf{V}$
\end{algorithmic} 
\end{algorithm}

We proposed a filtering threshold $th$ that eliminates samples that does not contain enough user data. The amount of valid transmission is represented by the sum of the total TB lengths in the window. We eliminated noise samples in the training and did not respond to samples that did not meet such condition during testing to avoid false alarms.

\subsection{Classification}

After constructing processing pipeline to convert sequence of physical channel records $(recs_i, y_i)_{i\in \{1,...,M\}}$ to numerical vectors $(\mathbf{V}_i, y_i)_{i\in\{1,...,M\}}$, the processed dataset was passed to machine learning models in the pursuit of accurate classification and fast computation. 

Linear Regression (LR) model and the Multi-Layer Perceptron (MLP) were evaluated in the first place as general baseline. Both are considered to be of least complexity, and thus has great advantage in efficiency considering existing low-end base station infrastructure. 

However, decision tree classifiers may be a better choice, as they often perform better with large datasets and are convenient to identify most influential factors for classification. The Classification and Regression Tree (CART) is one common approach, splitting at every node $N$ of the tree according to a threshold value on one feature. It uses the Gini impurity $Gini_N = 1 - \sum_{i=1}^{n}(p_{i,N})^2$ to approximate the cross entropy of each split, where $p_{i,N}$ is the proportion of samples classified to $i$th class by node $N$. 

Bagging and Boosting are two approaches of ensemble learning to inherit advantages of decision trees and achieve parallel computing in order to meet the requirements of real-time classification. Bagging can handle situations with insufficient data and unstable model performance and reduce variance. Random Forest (RF) combining $T$ CARTs $y_i^{pred,proba}=\frac{1}{T}\sum_{t=1}^Ttree_t(\mathbf{V}_i)$ was evaluated and compared with single CART to verify the improvement in accuracy. 

Given the objective of the binary classification task and the accessibility to large dataset, the boosting approach could be more proper for our task. It can effectively reduce the bias of the model, thereby coping with the huge feature size and finding a more appropriate high-dimensional mapping relationship. Taking Gradient Boosting Decision Tree (GBDT) for 0-1 binary classification as an example, one naive classifier $h_0(\cdot)$ is first constructed, performing random prediction without knowledge on any feature. In this case, the log odds for any sample in the train set $(\mathbf{V}_i, y_i)_{i\in\{1,...,M\}}$ are of the same value: 

\begin{equation}
 \log(odds)_i^0= \log\left(\frac{\sum_{j=1}^M\mathbbm{1}_{y_j=1}}{\sum_{j=1}^M\mathbbm{1}_{y_j=0}}\right) 
\end{equation}

\begin{figure}[htbp]
    \centering
    \includegraphics[width=\columnwidth]{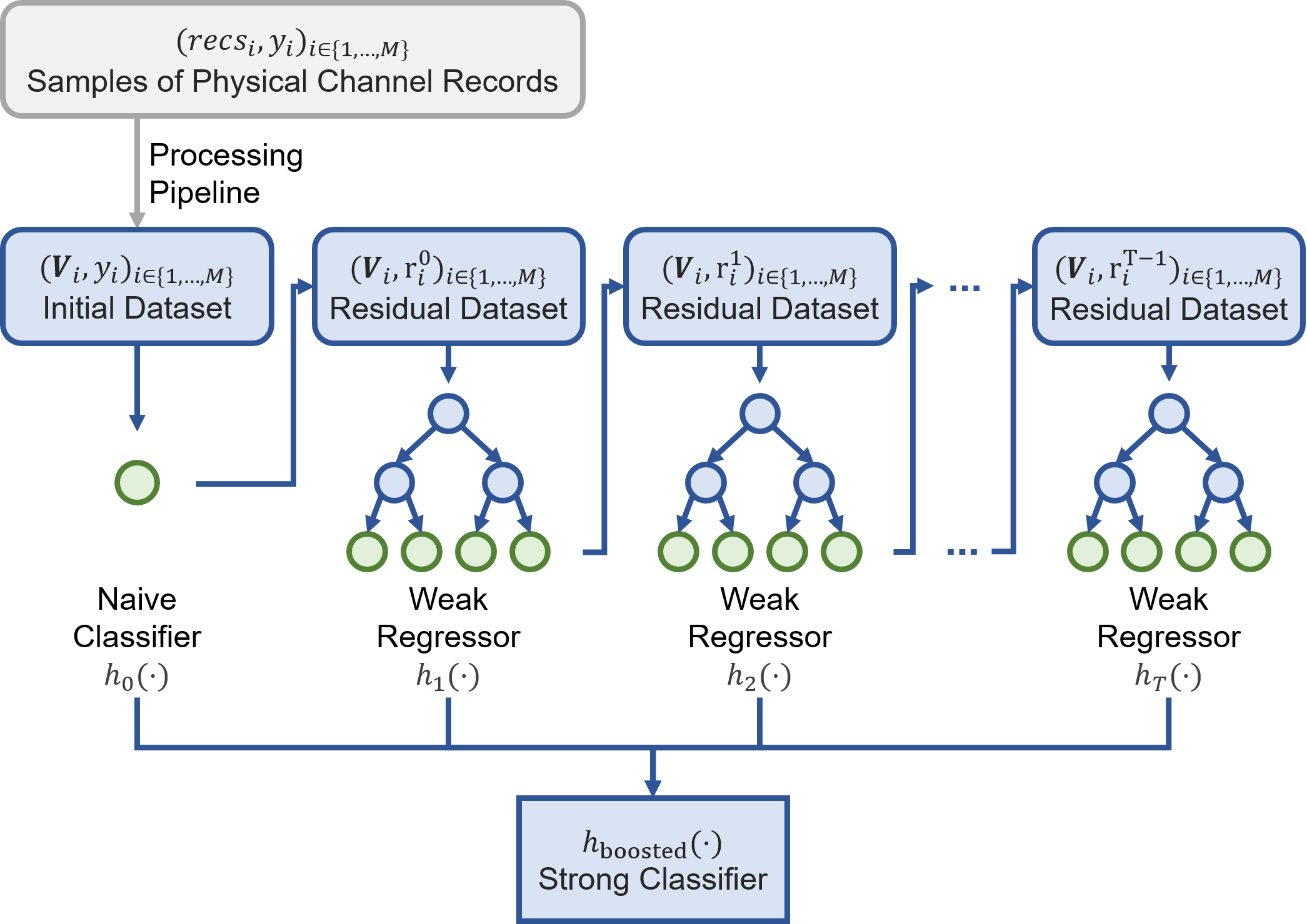}
    \caption{Decision-tree-based Gradient Boosting Framework}
    \label{fig:gb}
\end{figure}

After that, weak regressors can be trained sequentially, as illustrated in Figure \ref{fig:gb}. At every iteration $t\in\{1,...,T\}$, prediction probability and the residual $(p_i^{t-1}, r_i^{t-1})_{i\in\{1,...,M\}}$ are first calculated: 

\begin{align}
p_i^{t-1}:=& \mathbb{P}_{(h_\tau)_{\tau\in\{0,...,t-1\}}}(y_i=1) \\
=& \frac{\exp(\log(odds)_i^{t-1})}{1+\exp(\log(odds)_i^{t-1})}\\
r_i^{t-1}=&y_i-p_i^{t-1}
\end{align}

A new regression decision tree $h_t(\cdot)$ is trained on the residual train set $(\mathbf{V}_i, r_i^{t-1})_{i\in\{1,...,M\}}$. For leaf $L$ in $h_t$, the output of regression is calculated as follows, where $h_t(\mathbf{V}_i)\in L$ represents sample $i$ being classified into leaf $L$: 

\begin{equation}
output_L = \frac{\sum_{i=1}^Mr_i\mathbbm{1}_{h_t(\mathbf{V}_i)\in L}}{\sum_{i=1}^Mp_i(1-p_i)\mathbbm{1}_{h_t(\mathbf{V}_i)\in L}}
\end{equation}

And the overall log odds is accumulated for each sample $\log(odds)_i^t = \log(odds)_i^{t-1} + \gamma \cdot output_L |_{h_t(\mathbf{V}_i)\in L}$, with $\gamma$ the constant learning rate inferior to $1$ to avoid overfitting. 

After iterations, the classification prediction is extracted from the latest log odds, as an aggregation of all weak learners: 

\begin{equation}
h_{boosted}(\mathbf{V}_i) = \mathbbm{1}\left({\frac{\exp(\log(odds)_i^T)}{1+\exp(\log(odds)_i^T)} >0.5}\right)
\end{equation}

LGBM \cite{bib:lgb} is an optimised version of gradient boosting with CART. It uses histogram algorithm to replace the traditional sorting for splitting search and parallel computation. It also applies leaf-wise strategy with depth limitation to avoid low-yield splits. In addition, It implements Gradient-based One-side Sampling (GOSS) and Exclusive Feature Bunding (EFB) to reduce memory consumption and shorten training time. Extreme Gradient Boosting (XGB) \cite{bib:xgb} and CatBoost (CAT) \cite{bib:cat} are evaluated as two another implementations of GBDT, but have respective disadvantages. 

The performance of models was evaluated using classification metrics including the accuracy of prediction. However, many other factors such as training complexity, prediction response time and compatibility are important as well for application in industry production. 

\section{Experiments and Results}\label{sec:res}

In this section, learning frameworks were implemented in Python 3.9.16 on an AMD EPYC-7302 processor. The experiment was conducted on 170,000 records from different channels under two categories of traffic: website navigation and video streaming from YouTube. Hyperparameters were fixed with a window size $W=10\, ms$ and a filter threshold $th=150$ TBs per subframe, while others were tuned with cross validation grid search.

\begin{figure}[t!]
    \centering
    \includegraphics[width=\columnwidth]{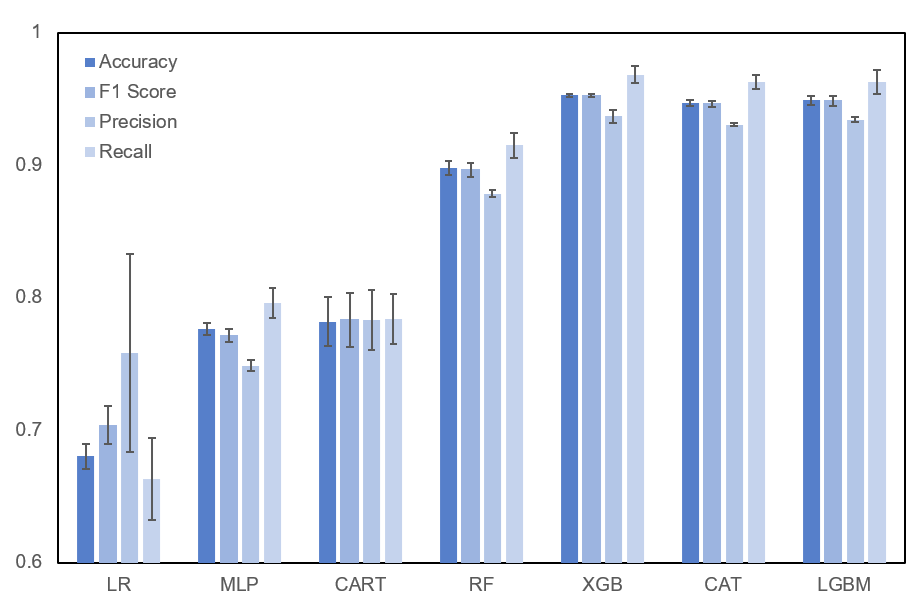}
    \caption{Comparison of Different ML Models With $W=10\, ms$ and $th=150$}
    \label{fig:comparison}
\end{figure}

The Figure \ref{fig:comparison} indicates a performance comparison of different models. We concluded that decision-tree-based gradient boosting algorithms outperforms baseline models and decision tree classifiers. LGBM along with XGB and CAT are able to achieve more than $95\%$ accuracy and good consistency. LGBM was chosen due to its training efficiency and the compatibility with related libraries. 

Besides, LGBM achieved prediction on more than 2000 test samples in less than a second. The size of the observation window is thus the main direct factor affecting the response time of the model.

\begin{figure}[b!]
    \centering
    \includegraphics[width=\columnwidth]{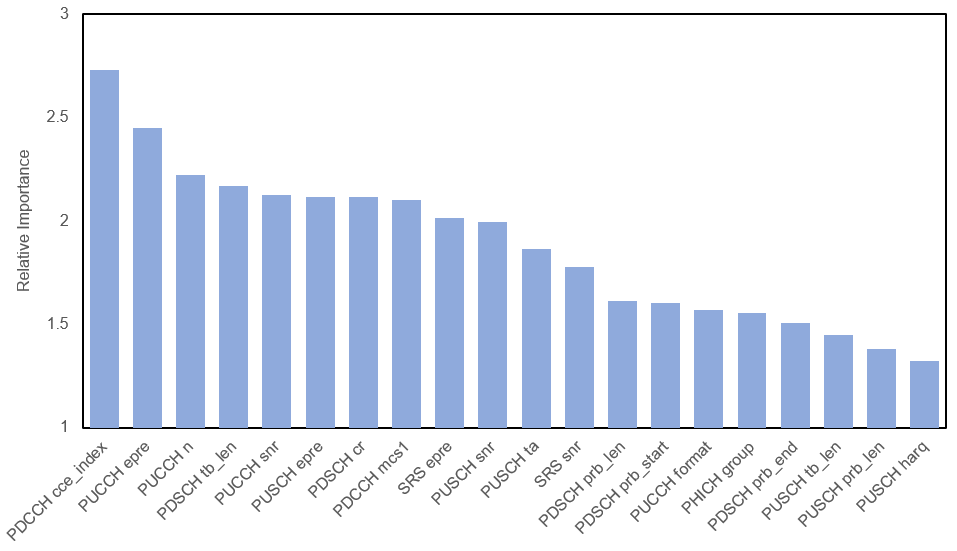}
    \caption{Relative Importance of Features in LGBM Classification With $W=10\, ms$ and $th=150$}
    \label{fig:importance}
\end{figure}

We further investigated the relative importance of features in the LGBM model according to their frequency of use for node partitioning. As revealed in Figure \ref{fig:importance}, the control channel element (CCE) index of the PDCCH channel and the EPRE of the PUCCH channel were of significant importance. This suggests that the position of the assigned RBs and the transmitted information density are correlated with traffic categories.

\begin{figure}[t!]
    \centering
    \includegraphics[height=0.67\columnwidth]{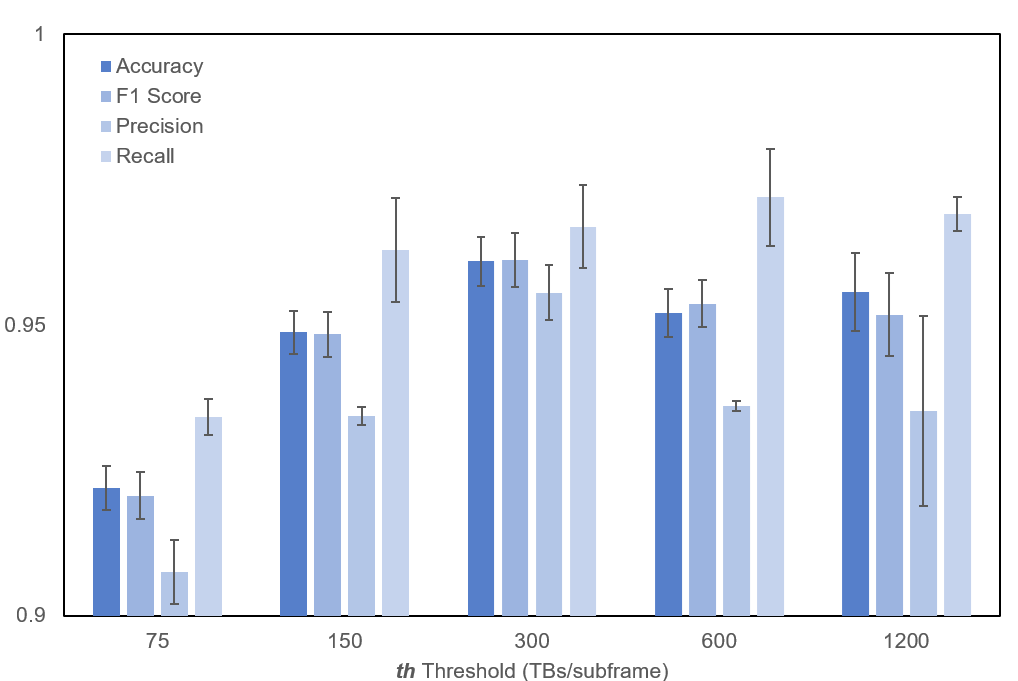}
    \caption{Influence of Noise Filtering $th$ on LGBM Classification, $W=10\, ms$}
    \label{fig:filtering}
\end{figure}

The importance of the noise filtering was also verified. The Figure \ref{fig:filtering} illustrates the improvement in accuracy from 92\% to 96\% by increasing the threshold $tb$ from 75 to 300 TBs per subframe. However, an excessively high threshold should be avoid as it may cause us to overlook samples with significant traffic flow characteristics but with lower throughput, thereby neglecting the rapid change in traffic. 

\begin{figure}[t!]
    \centering
    \includegraphics[height=0.67\columnwidth]{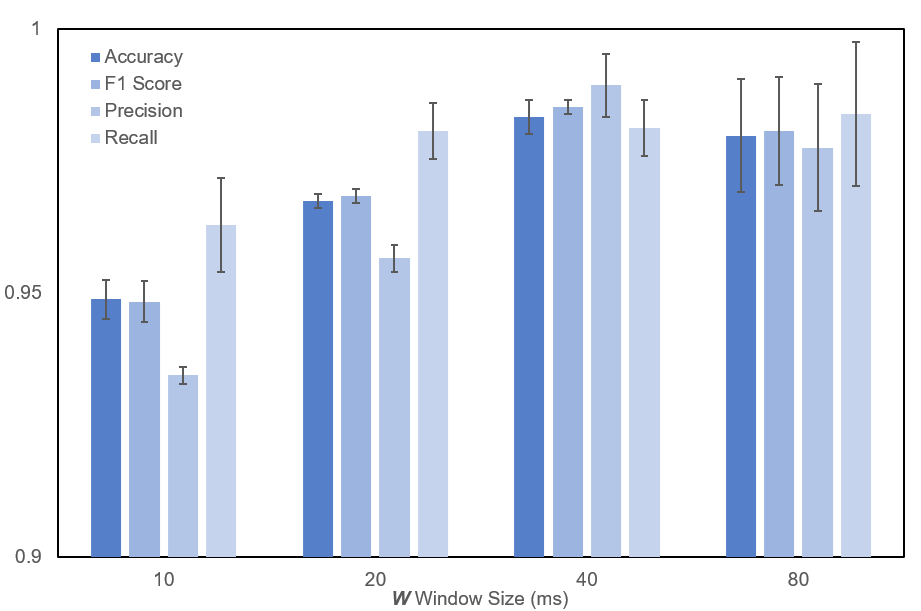}
    \caption{Influence of Window Size $W$ on LGBM Classification, $th=150$}
    \label{fig:window}
\end{figure}

Finally, we evaluated the effect of window size on the performance. Figure \ref{fig:window} demonstrates the successive improvement in performance by increasing the window size from $10 ms$ to $40 ms$. For window size larger than $40 ms$, the performance ceased to improve.

Overall, both hyperparameters represent a trade-off between traffic identification response speed and model performance. They should be adjusted according to the specific needs of downstream tasks. In our experiments, $W=10ms$ and $th=300$ TBs per subframe were found to be the choice without sacrificing our goals in real-time traffic classification. 

\section{Conclusion}\label{sec:con}

Utilizing physical channel features has been demonstrated to be an effective approach for extracting large amounts of information from very short time windows. The pipeline that converts physical channel record sequences into numerical vectors provides a widely applicable, high-performance interface for downstream tasks. Our experiments on a noise-limited 5G NSA dataset show that LGBM, a decision tree-based gradient boosting algorithm, can achieve over 95\% accuracy on classification tasks with a response time of 10 ms. We believe that the ease of training and parameter tuning, combined with the low computational cost of the LGBM model in practical application scenarios, makes it a highly suitable solution. Future research could involve exploring generalized scenarios with multiple mobile devices activated concurrently and with traffic from a broader range of applications. 

\end{document}